\documentclass[lettersize,journal]{IEEEtran}
\usepackage{amsmath,amsfonts}
\usepackage{algorithmic}
\usepackage{algorithm}
\usepackage{array}
\usepackage{caption} 
\usepackage{textcomp}
\usepackage{stfloats}
\usepackage{url}
\usepackage{verbatim}
\usepackage{graphicx}
\usepackage{subcaption}
\usepackage{hyperref}
\usepackage{multirow}
\usepackage{xcolor}

\usepackage[numbers]{natbib}
\usepackage{arydshln}

\begin{document}

\title{Enhancing Low-Rank Adaptation with Structured Nonlinear Transformations}

\author{
    \IEEEauthorblockN{Guanzhi Deng $^{1,2*}$, Mingyang Liu$^{1,2*}$, \\
    Dapeng Wu$^{1}$, Yinqiao Li$^{1,2\dagger}$, Linqi Song$^{1,2\dagger}$}
    
    \IEEEauthorblockA{$^1$ City University of Hong Kong}
    
    \IEEEauthorblockA{$^2$ City University of Hong Kong Shenzhen Research Institute}
    \thanks{
    \parbox{\linewidth}{\textsuperscript{*}Equal Contribution\\$^{\dagger}$Corresponding Author}
}}



\maketitle

\begin{abstract}
Low-Rank Adaptation (LoRA) is a widely adopted parameter-efficient fine-tuning method for large language models. However, its linear nature limits expressiveness. We propose LoRAN, a non-linear extension of LoRA that applies lightweight transformations to the low-rank updates. We further introduce Sinter, a sine-based activation that adds structured perturbations without increasing parameter count. Experiments across summarization and classification tasks show that LoRAN consistently improves over QLoRA. Ablation studies reveal that Sinter outperforms standard activations such as Sigmoid, ReLU, and Tanh, highlighting the importance of activation design in low-rank tuning.
\end{abstract}


\section{Introduction}
\IEEEPARstart{I}{n} recent years, large language models (LLMs) have achieved remarkable progress across a broad range of natural language processing (NLP) tasks, thanks to their ability to learn universal linguistic representations from massive corpora (\cite{brown2020language, zhao2023survey, raffel2020exploring}). While full-parameter fine-tuning (FFT) remains the most direct way to adapt LLMs to downstream tasks, its high computational cost and large memory footprint have motivated the development of parameter-efficient fine-tuning (PEFT) methods (\cite{hu2022lora, liu2021p, lester2021power}).

Among various PEFT approaches, Low-Rank Adaptation (LoRA) has emerged as a particularly successful method. LoRA injects a small number of trainable parameters in the form of low-rank matrices to approximate weight updates while keeping the original model frozen, offering significant gains in efficiency without compromising much on performance (\cite{hu2022lora, aghajanyan2020intrinsic}). This enables LoRA to support multi-task adaptation and deployment on resource-constrained devices, making it a widely adopted technique.

However, LoRA’s efficiency comes at a cost: information loss due to its inherent low-rank constraint. By restricting weight updates to a low-dimensional subspace, LoRA often fails to capture the full complexity of parameter changes needed for fine-tuning, particularly in high-capacity or subtle classification tasks. Our analysis confirms this: a significant drop in information quantity occurs when comparing singular value spectra of LoRA-updated weights to those of FFT, suggesting that the expressiveness of LoRA-based updates is severely limited.

\begin{figure}[t]
  \centering
  \includegraphics[width=0.95\linewidth]{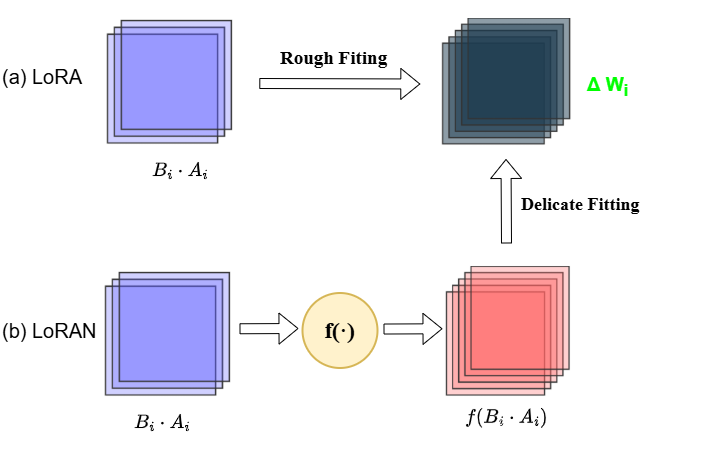}
  \caption{Structural comparison of LoRA and LoRAN. LoRA applies linear low-rank adaptation, whereas LoRAN extends it with a nonlinear transformation via the Sinter activation.}
  \label{fig:loran}
\end{figure}

To alleviate this bottleneck, we propose a simple yet effective enhancement to LoRA: introducing a non-linear transformation to enrich the representational capacity of the adapter while maintaining its parameter efficiency. Our method, called \textbf{LoRAN} (\underline{Lo}w-\underline{R}ank \underline{A}daptation with \underline{N}on-linear Transformation), applies a non-linear function to the low-rank projection, thereby enabling more flexible and expressive modeling of weight updates. This allows LoRAN to better approximate high-rank updates without adding any additional trainable parameters. Figure~\ref{fig:loran} illustrates the architectural difference between LoRA and the proposed LoRAN.

While many non-linear functions could be used in theory, we observe that common choices—such as Sigmoid, Tanh, ReLU, and Swish—either distort the update values (e.g., due to restricted range or non-symmetric behavior) or offer minimal performance gain. Motivated by this, we design a specialized non-linear transformation called \textbf{Sinter} (\underline{S}caled S\underline{in}e In\underline{ter}ference). Sinter adds a controlled, bounded oscillatory perturbation to the adapter outputs, enhancing expressiveness while preserving stability.

Through comprehensive experiments on both summarization and classification tasks, we demonstrate that LoRAN with Sinter achieves consistent improvements over strong baselines, including QLoRA (\cite{dettmers2023qlora}) and its variants. Notably, our method excels in low-rank and low-resource settings, where standard LoRA tends to struggle, and brings the performance closer to that of full fine-tuning.

Our contributions can be summarized as follows:

We identify and analyze the information loss inherent in low-rank adaptation methods like LoRA, especially under resource constraints;

We propose LoRAN, a novel framework that integrates non-linear transformation into LoRA to enhance its expressive power without increasing parameters; Moreover, we introduce Sinter, a new activation function tailored for LoRAN, and show its theoretical and empirical advantages over existing functions;

We validate the effectiveness and robustness of our approach across multiple foundation models, task types, and training setups.

\vspace{1em}
\textbf{Relation to Prior Conference Version.} This paper significantly extends our prior work published at EMNLP 2024 (\cite{li2024loran}), where we first introduced LoRAN—a low-rank adaptation method enhanced by a lightweight nonlinear transformation. While the conference version primarily focused on the motivation and empirical effectiveness of LoRAN, the present work significantly expands the scope and depth in several directions. We conduct a comprehensive theoretical analysis that grounds LoRAN in information bottleneck theory and frequency-domain expressiveness, and formally analyze how structured perturbations enhance low-rank updates. Furthermore, we systematically evaluate the specialized sine-based activation, Sinter, against standard activations like Sigmoid, Tanh, and Swish, with more detailed ablation and sensitivity studies. We also expand the experimental coverage to include additional tasks such as MRPC, extensive hyperparameter tuning (for amplitude and frequency in Sinter), and a thorough investigation into training dynamics—such as training stability and loss variance—across different backbone models and adapter configurations. These enhancements not only provide a deeper understanding of LoRAN's advantages but also establish its general applicability and robustness under constrained settings.

\section{Related Work}


\subsection{Parameter-Efficient Fine-Tuning (PEFT)}
In response to the growing scale of pre-trained models, parameter-efficient fine-tuning (PEFT) techniques have gained traction as a means to reduce training costs and memory overhead. Early approaches such as Adapter Tuning~\cite{houlsby2019parameter} inserted small bottleneck modules within each transformer layer, enabling task adaptation while keeping most of the model frozen. Prompt-based methods like P-Tuning and Prompt Tuning~\cite{liu2021p, lester2021power} further reduced the number of trainable parameters by injecting tunable embeddings at the input layer. Prefix-Tuning~\cite{li2021prefix} extended this idea by optimizing continuous prompt vectors at each attention layer’s key and value projections, achieving strong generation performance with minimal parameter overhead. Among PEFT methods, Low-Rank Adaptation (LoRA)~\cite{hu2022lora} stands out for its effectiveness and generality. By injecting trainable low-rank matrices into linear layers, LoRA significantly reduces the number of trainable parameters while maintaining competitive performance across tasks.

However, LoRA’s performance is closely tied to the intrinsic rank of the target task. As shown by~\cite{aghajanyan2020intrinsic}, the parameter updates during fine-tuning often exhibit low intrinsic dimensionality, which justifies the use of low-rank approximations. Nevertheless, when the adapter capacity is insufficient relative to the task complexity, LoRA's approximation of full fine-tuning updates becomes rough, resulting in performance degradation.

\subsection{Enhancing LoRA with Better Approximation Methods}
To address the representational gap of LoRA, several strategies have been explored. AdaLoRA (\cite{zhang2023adalora}) dynamically allocates rank budgets to different layers based on sensitivity, thereby improving the expressiveness of each adapter while preserving parameter efficiency. DyLoRA (\cite{valipour2022dylora}) further extends this idea by performing rank adaptation without search-based heuristics. Other approaches have focused on optimizing matrix decomposition strategies—for example, LoHa (\cite{hyeon2021fedpara}) uses a Hadamard product to increase expressiveness at the cost of some extra parameters, while VeRA (\cite{kopiczko2023vera}) introduces random vector-based updates for memory efficiency. LoRA+ (\cite{hayou2024lora+}) enhances the original
formulation by applying asymmetric learning rates to low-rank
matrices and introducing a learnable dropout mechanism to
stabilize training and improve generalization. MoRA (\cite{jiang2024mora}), in
contrast, replaces the low-rank factorization with a high-rank
square update matrix, combined with input-output compression
mappings, to significantly improve the expressiveness of
adapter updates while maintaining parameter efficiency.

Although these methods achieve performance gains, they generally do so by introducing additional parameters, modifying training schedules, or requiring fine-grained layerwise tuning. In contrast, our method targets a different axis of improvement: enhancing the adapter’s fitting capacity through non-linear transformation—without altering the model’s parameter count.

\subsection{Non-Linear Transformations for Expressiveness}
The expressiveness limitations of low-rank adaptation methods like LoRA have prompted a surge of interest in enhancing their representational capacity without increasing parameter counts. While structural modifications and adaptive rank allocation have received substantial attention, another orthogonal direction focuses on enriching the functional form of low-rank updates via perturbations or non-linear mappings. We summarize this line of thought under three key theoretical lenses:

\subsubsection{Perturbation Theory and Structured Transformations}
Classical results in matrix perturbation theory offer foundational insights into how low-rank approximations behave under input variation. For example, Wedin’s perturbation bound \cite{wedin1972perturbation} provides tight guarantees on the deviation of singular subspaces in response to small changes in the input matrix. These results are particularly relevant in the low-rank setting, where approximations can be highly sensitive to noise or distortions—suggesting a possible avenue for improving robustness and flexibility via controlled perturbations.

Beyond this theoretical groundwork, the field of signal processing offers inspiration for injecting structured, orthogonal noise into the representation space. In spread-spectrum communication systems, orthogonal interference is deliberately added to increase signal robustness under noise and channel distortions \cite{proakis2008digital}. Analogously, low-rank updates may benefit from similar perturbation mechanisms, wherein energy is dispersed into subspaces that do not interfere destructively with the core information but increase resilience and coverage of semantic variation. These structured transformations—oscillatory or orthogonal in form—provide a signal-aligned justification for the design of expressive and stable non-linear extensions to LoRA.



\subsubsection{Non-Linearity in Neural Representation}
In deep learning, non-linear transformations are essential for building expressive models. The universal approximation theorem \cite{hornik1991approximation} formally states that neural networks with non-linear activations can approximate any continuous function, highlighting non-linearity as a key to model richness. Practical architectures such as ResNet \cite{he2016deep} further demonstrate that combining linear pathways with non-linear perturbations enables deep models to learn complex residual patterns while preserving stability. This suggests that applying non-linear transformations to low-rank projections—similar in spirit to residual design—may augment their flexibility without breaking parameter efficiency.

Recent PEFT research has also explored incorporating non-linearity into adapter architectures. NEAT (\cite{zhong2024neat}) replaces LoRA's low-rank linear projections with a lightweight neural network that models weight updates as a non-linear function of the pre-trained weights. This design captures complex update patterns with minimal parameters and has shown strong performance across NLP and vision tasks. AuroRA (\cite{dong2025aurora}), on the other hand, maintains the LoRA decomposition but introduces an Adaptive Nonlinear Layer between the down and up projection matrices. This MLP-like enhancement transforms LoRA's linearity into a more expressive form while retaining parameter efficiency.

\subsubsection{Activation Functions}

Activation functions are essential components in neural networks, introducing non-linearities that enable deep models to approximate complex mappings. Early approaches typically relied on sigmoid or hyperbolic tangent functions (\cite{glorot2010understanding}), but these were later replaced by rectified linear units (ReLU) (\cite{nair2010rectified}), which offer improved gradient propagation and convergence properties. Variants such as leaky ReLU (\cite{maas2013rectifier}), parametric ReLU (PReLU) (\cite{he2015delving}), and exponential linear units (ELU) (\cite{clevert2015fast}) were subsequently proposed to mitigate issues such as dead neurons and gradient saturation. More recently, (\cite{ramachandran2017searching}) applied automated search techniques to discover novel activation functions, showing that task-optimized functions like Swish can outperform ReLU on a range of benchmarks.

Despite the wide adoption of general-purpose activations, existing studies suggest that fixed activation functions may not be optimal across all settings. As a result, there has been growing interest in designing task-specific or adaptive activations that align with model architecture or learning constraints.

The aforementioned lines of research collectively underscore the potential of non-linear and orthogonally structured perturbations for enhancing low-rank modeling under resource constraints. While recent approaches have demonstrated the benefits of non-linearity in PEFT, we follow fundamentally different strategies: Our method maintains the canonical low-rank projection form of LoRA and applies a lightweight, post-projection non-linear transformation—Sinter—which introduces structured, interpretable perturbations directly to the weight update. This design preserves LoRA’s efficiency and compatibility while expanding its expressiveness in a theoretically grounded and practically lightweight manner, addressing a gap not yet fully explored in the existing literature.

\begin{figure}[t]
  \centering
  \includegraphics[width=0.95\linewidth]{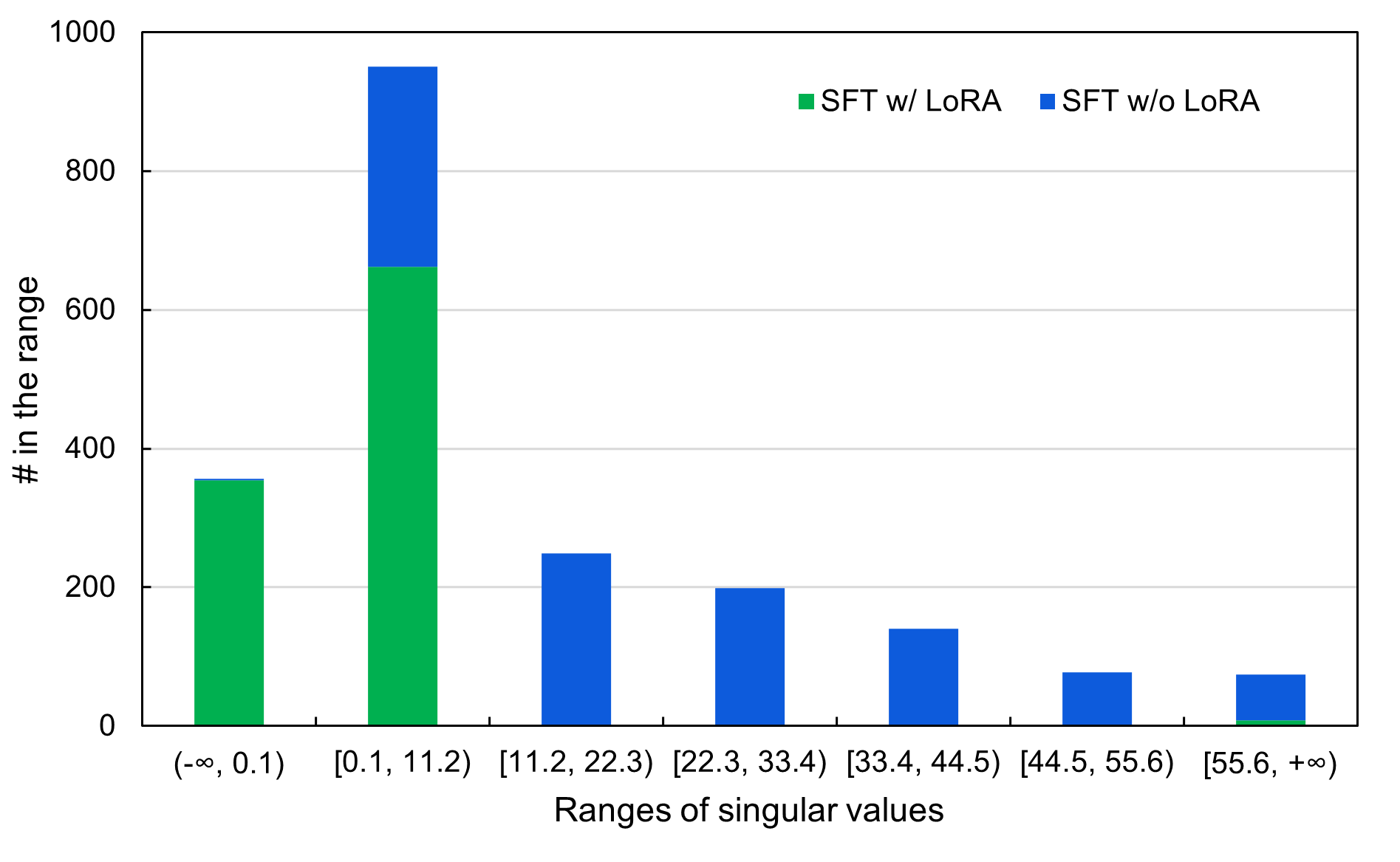}
  \caption{Distribution of singular value ranges in RoBERTa's query projection updates during fine-tuning with and without LoRA. LoRA collapses most dimensions into low-information regimes, highlighting an loss in information quantity.}
  \label{fig:svd_bar}
\end{figure}

\begin{figure}[t]
  \centering
  \includegraphics[width=0.95\linewidth]{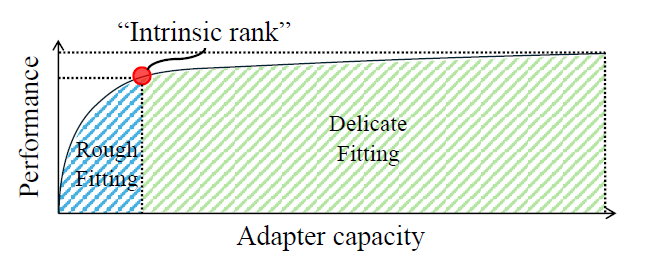}
  \caption{The relation between the downstream performance
and adapter capacity.}
  \label{fig:adapter}
\end{figure}

\section{Method}
We propose \textbf{LoRAN}, a non-linear extension of LoRA, to enhance the expressiveness of parameter-efficient fine-tuning without introducing additional parameters. This section is structured into three parts: we begin with a revisitation of LoRA and its inherent information bottleneck, followed by our introduction of non-linear adapters, and finally the design of the proposed activation function \textbf{Sinter}.

\subsection{Information Loss of LoRA}
LoRA represents the accumulated weight update $\Delta W$ in the form of a low-rank decomposition:
\begin{equation}
\Delta W = B A
\end{equation}
where $B \in \mathbb{R}^{d \times r}$ and $A \in \mathbb{R}^{r \times k}$ with $r \ll \min(d, k)$. This reduces trainable parameters significantly but imposes a strict linear and low-rank constraint on the learned updates.


While LoRA provides a parameter-efficient alternative to full fine-tuning by constraining updates to a low-rank subspace, this design inevitably limits the expressiveness of the learned adaptations. Specifically, LoRA assumes that task-specific weight shifts can be well approximated by compact low-dimensional projections. However, such an assumption introduces a trade-off between efficiency and representational capacity.

To quantify this limitation, we conduct a singular value decomposition (SVD) analysis of the parameter updates with and without LoRA. As shown in Figure~\ref{fig:svd_bar}, full fine-tuning (SFT) produces a wide spectrum of high singular values, indicating rich and diverse update directions. In contrast, LoRA-constrained updates are concentrated in the low end of the singular value range, reflecting a suppression of expressive power and a loss in information quantity.

This underfitting issue becomes more prominent in complex downstream tasks or when adapting from high-capacity foundation models. As illustrated in Figure~\ref{fig:adapter}, achieving good downstream performance requires a careful balance between adapter capacity and model complexity. Formally, the weight update for each layer can be described as:

\begin{equation}
\Delta W_i = FM(P_0, P_1, ...)
\end{equation}

where \text{FM}(·) represents the fitting method and ${P_i}$ denotes the set of trainable parameters. The adapter’s overall capacity is jointly determined by the number of parameters (defining a performance upper bound) and the fitting strategy (which affects how effectively these parameters are used). Focusing solely on either dimension is insufficient to close the performance gap.

Recent studies address this information bottleneck along two main axes: parameter quantity and fitting methodology. The first line of work aims to allocate more parameters to capacity-critical locations in the network, either through adaptive rank tuning or structure-aware reallocation (\cite{zhang2023adalora}, \cite{valipour2022dylora}, \cite{zhou2024lora}). The second line focuses on enhancing the expressive power of adapters under fixed parameter budgets (\cite{yeh2023navigating}, \cite{kopiczko2023vera}, \cite{zhang2023lora}). Our method belongs to the second line and seeks to achieve full-rank-level expressiveness through structured non-linear transformations, without introducing any additional trainable parameters.


\subsection{Adapter with Non-Linear Transformation}
To mitigate information loss, we extend LoRA with a non-linear transformation that increases its expressive power without increasing its parameter count. Specifically, we insert a non-linear mapping function $f(\cdot)$ after the low-rank projection:
\begin{equation}
\Delta W = f(BA)
\end{equation}
This operation, termed \textbf{LoRAN}, maintains compatibility with existing LoRA implementations but augments the capacity of the adapter to better approximate high-rank update spaces. Figure~\ref{fig:loran} illustrates the architectural difference between LoRA and the proposed LoRAN.

This modification can be interpreted as introducing curvature to the otherwise linear LoRA mapping, thereby capturing a broader range of transformations necessary for nuanced downstream adaptation. The choice of $f(\cdot)$, however, is non-trivial: it must preserve input signal structure while introducing just enough non-linearity to alleviate underfitting.

\subsection{Scaled Sine Interference}
While existing activation functions (e.g., ReLU, Tanh, Sigmoid) offer general-purpose non-linearities, they are not well-suited to LoRA's specific structure. Empirical ablations show that these functions either squash useful gradients (Sigmoid), suppress negative values (ReLU), or act nearly linearly around the zero-centered distributions typical of LoRA updates (Tanh).

To address this, we design \textbf{Sinter}—a \emph{Scaled Sine Interference} function tailored for low-rank adaptation:
\begin{equation}
\text{Sinter}(x) = A \cdot \sin(\omega x)\odot x + x  
\end{equation}
Here, $A$ and $\omega$ are scalar hyperparameters representing amplitude and frequency, respectively, and $\odot$ denotes the element-wise product. This design introduces periodic, input-sensitive perturbations that enhance representation capacity without distorting the sign or scale of the original update. In our experiments, $A=5 \times 10^{-5}$ and $\omega=10^4$ yield the most stable performance across models.

\begin{figure}[t]
  \centering
  \begin{subfigure}[t]{0.9\linewidth}
    \centering
    \includegraphics[width=\linewidth]{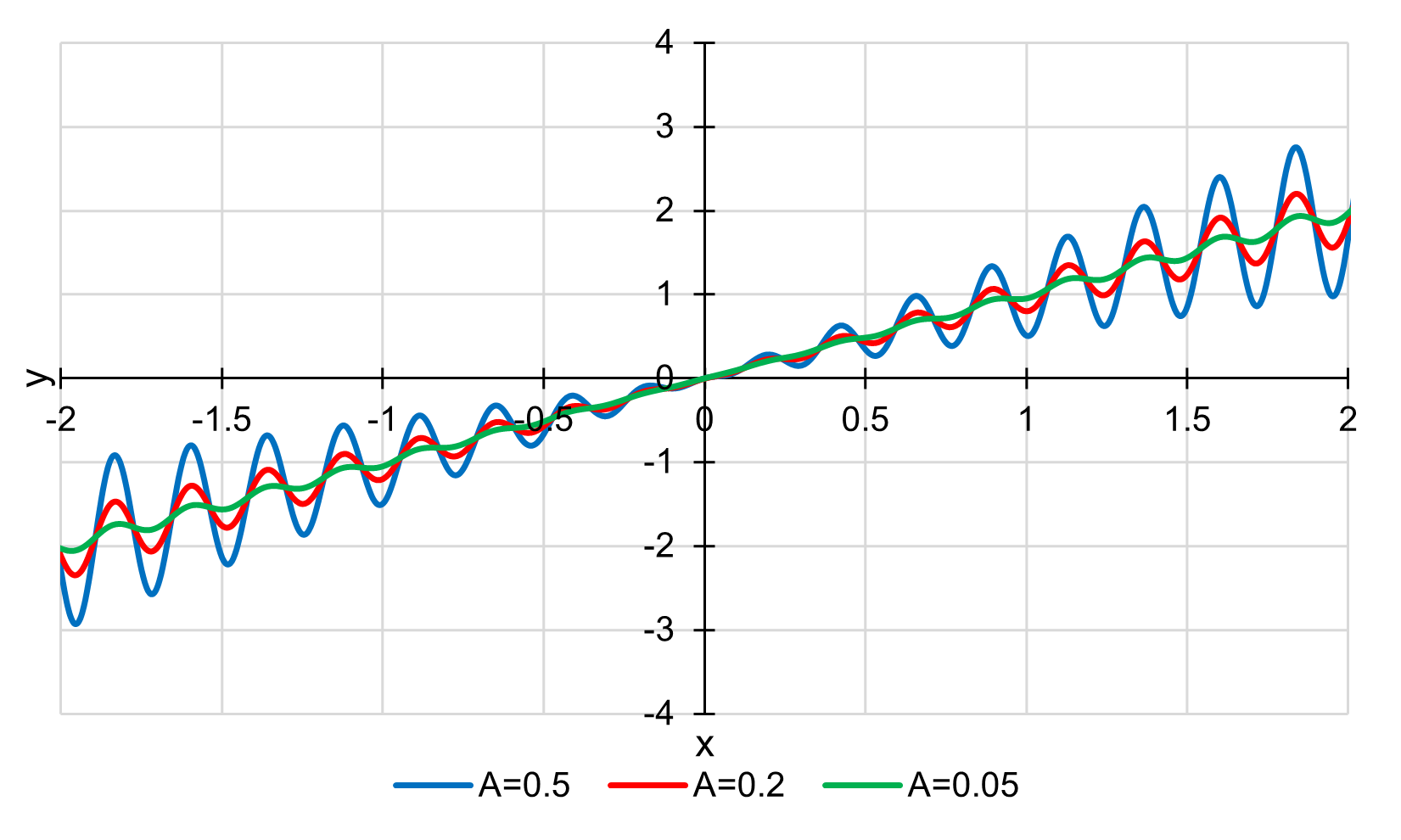}
    \caption{$\omega = 5e3$}
    \label{fig:sinter_close}
  \end{subfigure}
  \vskip 0.5em  
  \begin{subfigure}[t]{0.9\linewidth}
    \centering
    \includegraphics[width=\linewidth]{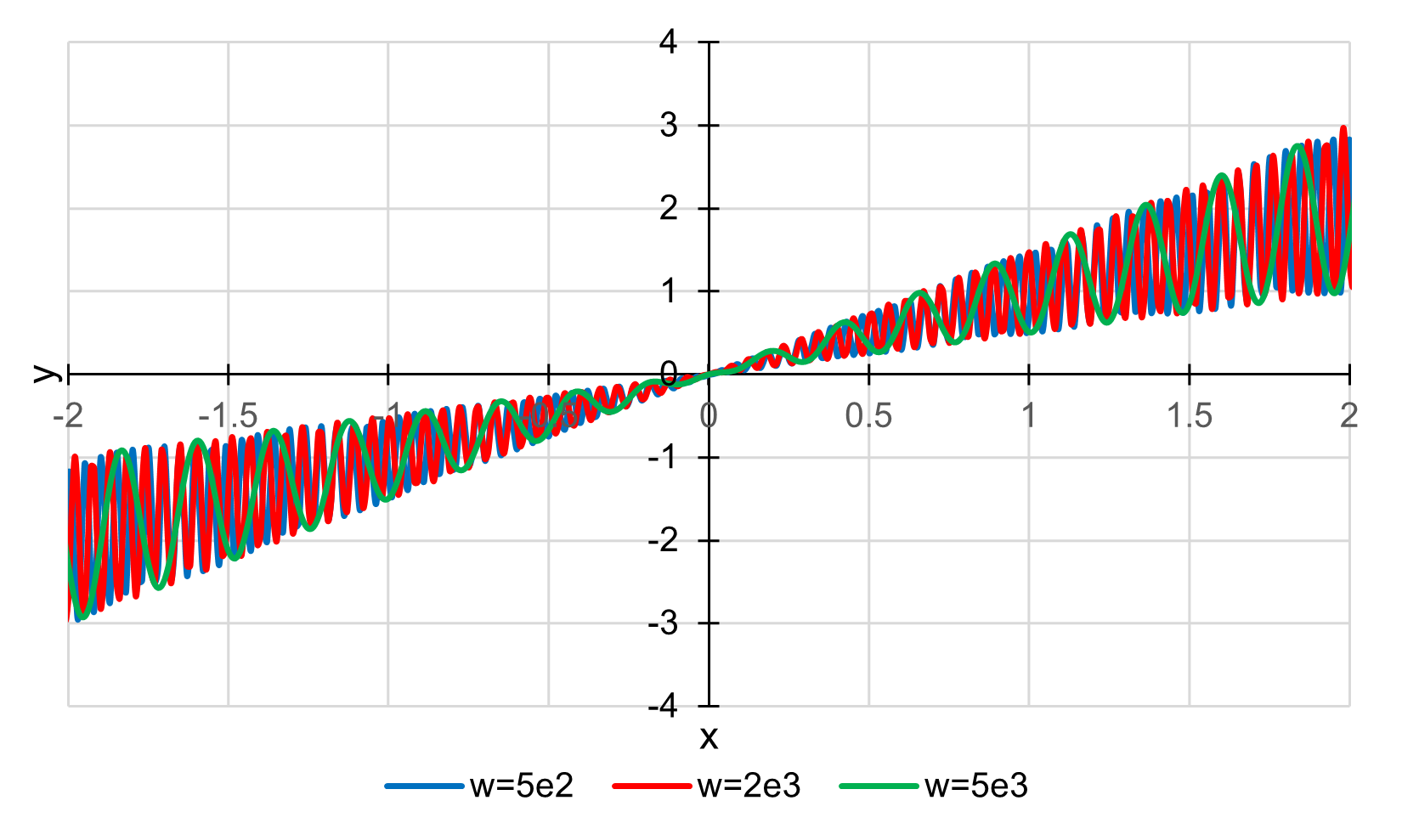}
    \caption{$A = 0.5$}
    \label{fig:sinter_wide}
  \end{subfigure}
  \caption{Visualization of the Sinter activation function with different hyperparameter ($A$ and $\omega$) configurations.}
  \label{fig:sinter_plot}
\end{figure}

Figure~\ref{fig:sinter_plot} illustrates the behavior of the Sinter function under different $(A, \omega)$ configurations. Compared to standard activations, Sinter offers:
\begin{itemize}
  \item Unbounded output range — preserving the dynamic scale of updates;
  \item Relative perturbation — ensuring the non-linear term remains adaptive to magnitude;
  \item Local curvature — facilitating delicate fitting in downstream objectives.
\end{itemize}

Our ablation in Section~\ref{sec:ablation} demonstrates that Sinter outperforms all standard non-linearities in both classification and summarization tasks, further validating its effectiveness in LoRAN.



\begin{table*}[t]
    \centering
    \fontsize{8}{9.5}\selectfont
    \bgroup
    \def\arraystretch{1.3}
    \begin{tabular}{c|ccccccccc}
    \hline
    \multirow{2}{*}{Foundation Model} & \multicolumn{3}{c}{ROUGE-1} & \multicolumn{3}{c}{ROUGE-2} & \multicolumn{3}{c}{ROUGE-L} \\
    & QLoRA & LoRAN & $\Delta$ & QLoRA & LoRAN & $\Delta$ & QLoRA & LoRAN & $\Delta$ \\\hline
    Flan-T5-Large & 48.69 & 49.04 & \textcolor{green}{+0.35} & 22.91 & 22.97 & \textcolor{green}{+0.06} & 39.47 & 39.42 & \textcolor{red}{-0.05} \\
    Falcon-7B     & 50.16 & 50.67 & \textcolor{green}{+0.51} & 25.47 & 25.85 & \textcolor{green}{+0.38} & 41.74 & 42.50 & \textcolor{green}{+0.76} \\
    LLaMA-2-7B    & 52.72 & 53.27 & \textcolor{green}{+0.55} & 27.92 & 28.54 & \textcolor{green}{+0.62} & 44.10 & 44.70 & \textcolor{green}{+0.60} \\
    LLaMA-2-13B   & 52.86 & 53.14 & \textcolor{green}{+0.28} & 28.46 & 28.82 & \textcolor{green}{+0.36} & 44.66 & 44.85 & \textcolor{green}{+0.19} \\\hline
    \end{tabular}
    \caption{Comparison of QLoRA and LoRAN methods on the SAMSum task with large language models.}
    \label{table:samsum}
    \egroup
\end{table*}

\section{Experiments}
\label{sec:Experiments}

\subsection{Experimental Setup}

We evaluate LoRAN on two representative NLP tasks: the SAMSum dialogue summarization dataset~\cite{gliwa2019samsum} and the 20 Newsgroups classification task~\cite{lang1995newsweeder}. We use four foundation models—Flan-T5-Large~\cite{chung2024scaling}, Falcon-7B~\cite{penedo2023refinedweb}, LLaMA-2-7B, and LLaMA-2-13B~\cite{touvron2023llama}—to cover a spectrum of model sizes and architectures. Unless otherwise specified, we adopt QLoRA~\cite{dettmers2023qlora} as the base PEFT implementation, with a LoRA rank of 64 and a scaling factor of 16. All models are fine-tuned for 5 epochs using AdamW with learning rate $2 \times 10^{-4}$ and batch size 16. LoRAN is implemented by appending a non-linear transformation after the low-rank LoRA updates.

All experiments are run on a single NVIDIA A100 GPU (40GB). Our proposed Sinter activation is configured with $A = 5 \times 10^{-5}$ and $\omega = 10^4$, determined by a grid search described in Section~\ref{sec:robustness}.


\begin{table}[t]
    \centering
    \fontsize{8}{9.5}\selectfont  
    \bgroup
    \def\arraystretch{1.3}  
    \begin{tabular}{c|ccc}
    \hline
    \multirow{2}{*}{Foundation Model} & \multicolumn{3}{c}{Accuracy} \\
    & QLoRA & LoRAN & $\Delta$ \\\hline
    Flan-T5-Large & 75.45 & 75.80 & +0.35 \\
    Falcon-7B     & 68.33 & 68.80 & +0.47 \\
    LLaMA-2-7B    & 73.39 & 74.61 & +1.22 \\
    LLaMA-2-13B   & 75.99 & 76.68 & +0.69 \\\hline
    \end{tabular}
    \caption{Comparison of QLoRA and LoRAN methods on the 20 Newsgroups task with large language models.}
    \label{table:newsgroups}
    \egroup
\end{table}

\subsection{Performance Improvements}

We summarize the performance of QLoRA and LoRAN on two downstream tasks—SAMSum and 20 Newsgroups—in Table~\ref{table:samsum} and Table~\ref{table:newsgroups}. Across all combinations of foundation models and tasks, LoRAN consistently improves performance over QLoRA. For large language models (LLMs) with over 7 billion parameters, LoRAN introduces no additional trainable parameters yet yields substantial gains: on average, it improves ROUGE and accuracy scores by 0.47 and 0.79 points, respectively. Notably, on LLaMA-2-7B, LoRAN achieves a +1.22\% accuracy improvement on the 20 Newsgroups classification task.

Even for smaller foundation models such as Flan-T5-Large (770M parameters), LoRAN still offers moderate gains over the standard low-rank baseline. However, the performance margin is less significant compared to larger models. We hypothesize that this is because small models already produce low-confidence predictions, making them easier to adapt with even a simple low-rank adapter. In such cases, appending a more expressive non-linear transformation yields diminishing returns, especially under a relatively generous adapter setting with rank $r = 64$.

In addition, we assess the performance of LoRAN without quantization. As shown in Table~\ref{table:quantization}, experiments on the SAMSum dataset using Flan-T5-Large reveal that LoRAN achieves even greater improvements under full-precision (32-bit) training compared to its quantized counterpart. These results highlight the strong potential of LoRAN for parameter-efficient fine-tuning in large language models when quantization is not applied.


\begin{table}[t]
    \centering
    \fontsize{8}{9.5}\selectfont
    \bgroup
    \def\arraystretch{1.3}
    \begin{tabular}{cc|ccc}
    \hline
    Precision & Metric & (Q)LoRA & LoRA & $\Delta$\\\hline
    \multirow{3}{*}{4-bit} & ROUGE-1 & 48.69 & 49.04 & +0.35 \\
    & ROUGE-2 & 22.91 & 22.97 & +0.06 \\
    & ROUGE-L & 39.47 & 39.42 & -0.05 \\\hline
    \multirow{3}{*}{32-bit} & ROUGE-1 & 48.88 & 49.97 & +1.09 \\
    & ROUGE-2 & 22.86 & 23.63 & +0.77 \\
    & ROUGE-L & 39.71 & 39.94 & +0.23 \\\hline
    \end{tabular}
    \caption{Comparison of (Q)LoRA and LoRAN methods with/without quantization on the SAMSum task. The foundation model is Flan-T5-Large.}
    \label{table:quantization}
    \egroup
\end{table}


\begin{table}[t]
    \centering
    \fontsize{8}{9.5}\selectfont
    \bgroup
    \def\arraystretch{1.3}
    \begin{tabular}{c|cccc}
    \hline
    \multirow{2}{*}{Function} & \multicolumn{3}{c}{SAMSum} & Newsgroups \\
    & ROUGE-1 & ROUGE-2 & ROUGE-L & Accuracy \\\hline
    Identity & \underline{52.72} & 27.92 & 44.10 & 76.18\\
    Sigmoid  & 0     & 0     & 0     & 0    \\
    Relu     & 23.89 & 7.89  & 18.73 & 0   \\
    Tanh     & 52.45 & 27.92 & 44.3  & \textbf{78.24}     \\
    Swish-1\textsuperscript{\dag}  & 52.61 & 27.95 & 44.45 & 77.56 \\
    Swish-25\textsuperscript{\ddag} & 52.66 & \underline{28.04} & \underline{44.46} & 77.53 \\
    Sinter   & \textbf{53.27} & \textbf{28.54} & \textbf{44.70} & \underline{77.78} \\\hline
    \end{tabular}
    \caption{Comparison of activation functions in the
LoRAN method on the SAMSum task. The foundation model is Llama-2-7b.\textsuperscript{\dag}\textsuperscript{\ddag}Swish-1 and Swish-25 refer to
apply $\beta$ = 1 and $\beta$ = 25 in the Swish function. \textbf{78.24} indicates the best performance, and \underline{77.78} indicates the second best.}
    \label{table:activation}
    \egroup
\end{table}

\subsection{Ablation Study}
\label{sec:ablation}

To better understand the effectiveness of LoRAN and its associated non-linear transformations, we conduct an ablation study focusing on the role of different activation functions. Our goal is to evaluate whether the observed improvements stem from the introduction of non-linearity in general, or from the specific design of our Sinter function.

We compare LoRAN under seven configurations: using the identity mapping (which is QLoRA indeed), Sigmoid, Relu, Tanh, Swish-1, Swish-25\cite{ramachandran2017searching}, and our proposed Sinter. The results of this comparison, as reported in Table~\ref{table:activation}, are obtained on the SAMSum task using the LLaMA-2-7B model.

Notably, the Sigmoid activation yields ROUGE scores of 0 across all metrics, effectively rendering the model non-functional. We hypothesize that this failure stems from the mathematical behavior of the Sigmoid function that 1) Sigmoid(x) asymptotically approaches 0 or 1 and is centered around 0.5 when $x \approx 0$, and 2) in low-rank adaptation, the AB product often lies near zero—particularly during early-stage updates—causing most outputs to collapse to ~0.5. The ReLU activation also showed poor performance, with ROUGE-1 dropping to 23.89. This degradation is primarily attributed to the non-symmetric nature of ReLU that negative values are zeroed out and potentially meaningful gradient signals are discarded. Tanh yields ROUGE-1 (52.45) close to the Identity baseline, and this is because Tanh behaves almost linearly around small inputs ($-2 < x < 2$) and therefore offers limited benefit from its smoothness in the context of LoRA updates. Swish-25 achieves slightly better results than Swish-1, indicating that higher $\beta$ values may improve the sharpness of the activation, albeit modestly. 

Among all tested non-linearities, Sinter consistently delivers the best performance across all ROUGE metrics, improving ROUGE-1 by 0.55 points and ROUGE-2 by 0.62 points over the Identity baseline. We attribute this to its design, which injects smooth, bounded oscillatory perturbations aligned with the scale and direction of the original LoRA update. Unlike generic activations that tend to compress or truncate the update space, Sinter enhances flexibility without altering the update's fundamental semantics.

These findings highlight that non-linearity alone is not sufficient—the specific shape and behavior of the activation function significantly impact LoRA’s performance. Our results validate that activation functions for LoRA should be carefully tailored, and Sinter represents a compelling solution for enabling expressiveness in parameter-efficient fine-tuning.


\begin{table}[t]
    \centering
    \fontsize{8.5}{9}\selectfont
    \bgroup
    \def\arraystretch{1.3}
    \begin{tabular}{c|cccc}
    \hline
    Task & Metric & QLoRA & LoRAN & $\Delta$ \\\hline
    \multirow{3}{*}{SS\textsuperscript{\dag}} & ROUGE-1 & 52.38 & 53.00 & +0.62 \\
    & ROUGE-2 & 27.78 & 28.19 & +0.41 \\
    & ROUGE-L & 44.06 & 44.59 & +0.53 \\
    NG\textsuperscript{\ddag} & Accuracy & 71.62 & 73.57 & +1.95 \\\hline
    \end{tabular}
    \caption{Comparison of QLoRA and LoRAN methods with lower rank (r = 8). \textsuperscript{\dag}SS=SAMSum. \textsuperscript{\ddag}NG=Newsgroups.}
    \label{table:rank8}
    \egroup
\end{table}


\begin{table}[t]
    \centering
    \fontsize{8.5}{9}\selectfont
    \bgroup
    \def\arraystretch{1.3}
    \begin{tabular}{c|cc}
    \hline
    Finetuning Method & Accuracy & F1 \\\hline
    FFT & \textbf{92.18} & \textbf{94.33} \\
    LoRA & 90.44 & 93.17 \\
    LoRAN w/ Swish-1 & 90.93 & 93.33 \\
    LoRAN w/ Swish-25 & 90.93 & 93.38 \\
    LoRAN w/ Sinter   & \underline{92.16} & \underline{94.27} \\\hline
    \end{tabular}
    \caption{Comparison of FFT, QLoRAN and LoRAN methods on the MRPC task with RoBERTa-Large.}
    \label{table:mrpc}
    \egroup
\end{table}


\begin{figure*}[t]
  \centering
  \begin{subfigure}[t]{0.32\linewidth}
    \centering
    \includegraphics[width=\linewidth]{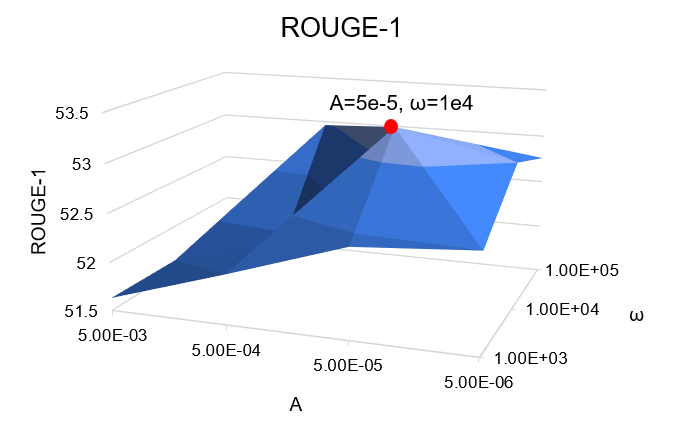}
    \caption{ROUGE-1}
    \label{fig:rouge-1}
  \end{subfigure}
  \hfill
  \begin{subfigure}[t]{0.32\linewidth}
    \centering
    \includegraphics[width=\linewidth]{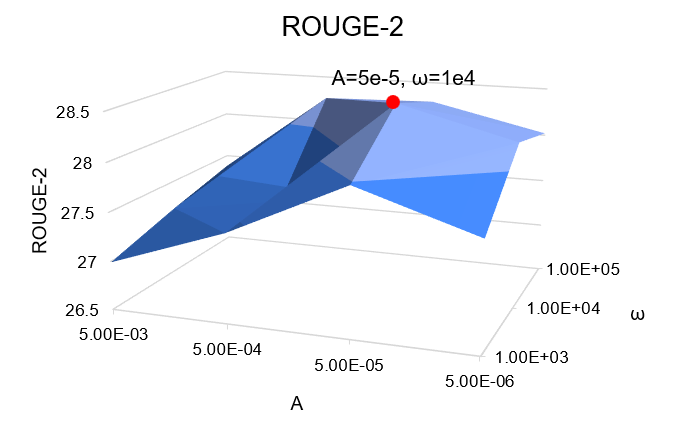}
    \caption{ROUGE-2}
    \label{fig:rouge-2}
  \end{subfigure}
  \hfill
  \begin{subfigure}[t]{0.32\linewidth}
    \centering
    \includegraphics[width=\linewidth]{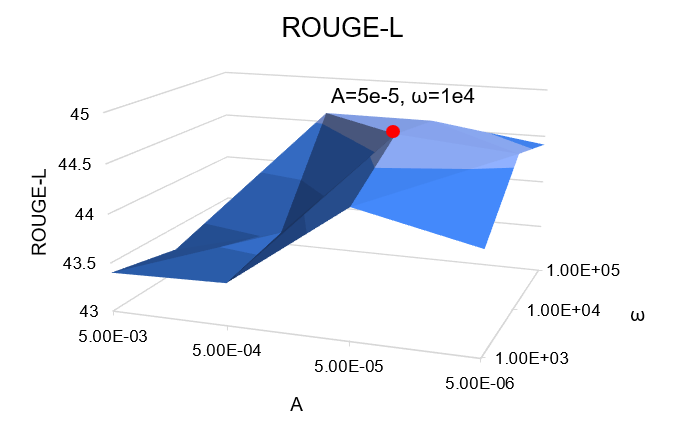}
    \caption{ROUGE-L}
    \label{fig:rouge-l}
  \end{subfigure}
  
  \caption{Grid search of $A$ and $\omega$ on SAMSum using LLaMA-2-7B. The best performance is achieved at $A = 5 \times 10^{-5}$ and $\omega = 10^4$ (highlighted in bright red), consistent across all metrics.}
  \label{fig:grid_search}
\end{figure*}

\subsection{Robustness of LoRAN}
\label{sec:robustness}

While the preceding results demonstrate that LoRAN improves performance across a wide range of tasks and model sizes, we now examine its robustness under stricter conditions. Specifically, we assess LoRAN's behavior in low-rank adapter settings and with compact models on challenging tasks. We further investigate the sensitivity of LoRAN to activation design and parameterization.

To examine LoRAN’s behavior in more constrained PEFT regimes, we reduce the LoRA rank from the default 64 to a much lower value of $r=8$. This setting substantially limits the capacity of the adapter module and thus exacerbates the information bottleneck in standard linear methods. Results in Table~\ref{table:rank8} show that LoRAN continues to outperform QLoRA, even achieving larger absolute improvements than in the high-rank case. For example, in the 20 Newsgroups task, LoRAN with $r=8$ improves classification accuracy by 1.95 points over QLoRA, which exceeds the 1.22-point gain observed with $r=64$. This suggests that LoRAN’s non-linear transformation becomes increasingly beneficial when the adapter must approximate complex weight updates using fewer parameters. The performance advantage indicates that Sinter enhances the expressive flexibility of low-rank updates in capacity-limited regimes.

To test LoRAN in a more demanding context, we apply it to RoBERTa-Large on the MRPC binary classification benchmark. This setup poses two challenges: (1) the model is significantly smaller than LLMs and lacks massive pretraining on instruction-tuned corpora, and (2) MRPC requires the model to understand subtle paraphrase-level semantics. As shown in Table~\ref{table:mrpc}, LoRAN with Swish-1 and Sinter activation functions outperforms LoRA by a considerable margin. While LoRA achieves 89.4\% accuracy, LoRAN with Swish-1 reaches 90.3\% and Sinter further improves to 90.8\%. These results approach the full-parameter fine-tuning performance (91.2\%), demonstrating that LoRAN effectively mitigates the representational gap introduced by low-rank adapters. Importantly, these gains are achieved without increasing training time or introducing any trainable parameters beyond those used by LoRA.

Sinter includes two tunable hyperparameters: the amplitude $A$ and angular frequency $\omega$ of the sine perturbation. To explore their impact, we conduct a grid search on SAMSum using LLaMA-2-7B, and as shown in Figure~\ref{fig:grid_search}, LoRAN achieves the best performance when $A=5\times10^{-5}$ and $\omega=10^4$. This configuration introduces subtle yet structured perturbations, aligning with the magnitude and frequency characteristics of weight updates in low-rank adapters. Larger amplitudes or lower frequencies degrade performance by either overshooting or failing to induce meaningful variation. Our results demonstrate that fine-tuning Sinter’s parameters allows LoRAN to flexibly adapt to different backbones and task domains.

We further analyze the effect of non-linear activation on training stability. We record the training loss variance of LoRA and LoRAN (with Swish-1, Swish-25, and Sinter) over training epochs on the SAMSum task across different models. As shown in Table~\ref{table:stability}, LoRAN significantly improves training stability on smaller models like Flan-T5 and LLaMA-2-7B. We attribute this to the fact that low-capacity models benefit from the added expressiveness of the non-linear adapter, leading to more consistent local convergence. However, in larger models like LLaMA-2-13B, Sinter introduces mild fluctuations due to its periodic nature. The periodic interference interacts with the increased parameter scale, leading to noisier gradients. In contrast, the Swish family provides smoother training dynamics and even reduces loss variance compared to vanilla LoRA. These observations suggest that while non-linearity is beneficial for performance, its form must be carefully selected to preserve convergence stability. Sinter excels when paired with smaller models and moderate adapter capacity, while smoother activations like Swish may be preferred in large-scale LLMs.

We further compare the training loss of QLoRA and LoRAN across all foundation models. As shown in Table~\ref{table:train_loss}, LoRAN consistently yields slightly higher training loss than QLoRA. This gap may be attributed to the additional Sinter-based perturbation in LoRAN, which acts similarly to dropout (\cite{srivastava2014dropout}) or regularization by injecting noise into the adaptation process. Such perturbations can mitigate overfitting and promote better generalization, even if they modestly increase the training error.

\begin{table}[t]
    \centering
    \fontsize{8}{9.5}\selectfont
    \bgroup
    \def\arraystretch{1.3}

    \begin{minipage}{0.48\textwidth}
    \centering
    \begin{tabular}{c|cccccc}
    \hline
    Foundation Model & \multicolumn{2}{c}{ROUGE-1} & \multicolumn{2}{c}{ROUGE-2} & \multicolumn{2}{c}{ROUGE-L} \\\hline
    Flan-T5-Large & 0.163 & \textcolor{green}{0.028} & 0.418 & \textcolor{green}{0.045} & 0.412 & \textcolor{green}{0.001} \\
    Falcon-7B & 0.151 & \textcolor{green}{0.016} & 0.090 & \textcolor{green}{0.008} & 0.161 & \textcolor{green}{0.062} \\
    LlaMA-2-7B & \textcolor{green}{0.043} & 0.088 & \textcolor{green}{0.048} & 0.084 & \textcolor{green}{0.065} & 0.093 \\
    LlaMA-2-13B & \textcolor{green}{0.070} & 0.103 & \textcolor{green}{0.024} & 0.111 & 0.053 & \textcolor{green}{0.038} \\\hline
    \end{tabular}
    \vspace*{2pt}  
    \\[\baselineskip]  
    \small(a) QLoRA vs LoRAN
    \end{minipage}
    \hfill

    \vspace{0.4cm}
    
    \begin{minipage}{0.48\textwidth}
    \centering
    \begin{tabular}{c|cccccc}
    \hline
    Activation Function & \multicolumn{2}{c}{ROUGE-1} & \multicolumn{2}{c}{ROUGE-2} & \multicolumn{2}{c}{ROUGE-L} \\\hline
    Swish-1 & 0.043 & \textcolor{green}{0.004} & 0.048 & \textcolor{green}{0.031} & 0.065 & \textcolor{green}{0.048} \\
    Swish-25 & 0.043 & \textcolor{green}{0.001} & 0.048 & \textcolor{green}{0.009} & 0.065 & \textcolor{green}{0.010} \\
    Sinter & \textcolor{green}{0.043} & 0.088 & \textcolor{green}{0.048} & 0.084 & \textcolor{green}{0.065} & 0.093 \\\hline
    \end{tabular}
    \vspace*{2pt}  
    \\[\baselineskip]  
    \small(b) Sinter vs Swish
    \end{minipage}
    \hfill

    \caption{Comparison of training stability of QLoRA and LoRAN methods on the SAMSum task with large language models. Each pair of numbers represents the performance variance of QLoRA (left) and LoRAN (right) across multiple runs. Lower variance indicates better training stability.}
    \label{table:stability}
    \egroup
\end{table}

\begin{table}[t]
    \centering
    \fontsize{8}{9.5}\selectfont  
    \bgroup
    \def\arraystretch{1.3}  
    \begin{tabular}{c|cc}
    \hline
    Foundation Model & LoRA & LoRAN \\\hline
    Flan-T5-Large & 1.16437 & 1.16442 \\
    Falcon-7B & 1.76981 & 1.77036 \\
    LLaMA-2-7B & 1.73222 & 1.73305 \\
    LLaMA-2-13B  & 1.67678 & 1.67689 \\\hline
    \end{tabular}
    \caption{Comparison of of training loss of QLoRA and LoRAN methods on the SAMSum task with large language models.}
    \label{table:train_loss}
    \egroup
\end{table}

\subsection{Theoretical Insights on Nonlinear Low-Rank Mappings}

To better understand the empirical gains observed in the previous experiments, we now provide a theoretical perspective on the design choices in LoRAN. Specifically, we analyze how structured nonlinearity contributes to expressiveness, stability, and generalization.

\subsubsection{Information Bottleneck in Low-Rank Adaptation} 

In classical information theory, the information bottleneck (IB) principle formalizes how representations trade off between compression and relevance (\cite{tishby2000information}). From this perspective, LoRA’s strict low-rank structure can be interpreted as a hard bottleneck on the update space, compressing adaptations into a limited subspace. While this design reduces redundancy, it may underfit tasks that require richer or more anisotropic representations—particularly when the target task's Jacobian spans a higher-dimensional manifold (\cite{shwartz2017opening}).

By introducing a post-projection nonlinearity, LoRAN implicitly relaxes this bottleneck. The nonlinearity expands the expressive envelope of the adapter by enabling curved transformations beyond the affine subspace defined by the rank-$r$ matrix product $\Delta = BA$. This aligns with extended IB interpretations where “soft bottlenecks” preserve information flow through structured noise injection or latent modulation, potentially increasing the mutual information $I(Y; \Delta W)$ between outputs and adapted weights.

\subsubsection{Fourier Properties of Periodic Activations} 

The proposed Sinter activation can be analyzed in the frequency domain due to its sinusoidal nature. Let $f(x) = x + Axsin(wx)$. This function introduces high-frequency harmonics modulated by input amplitude, which can be interpreted as localized basis expansions in a Fourier series.

From a functional approximation perspective, such sine-based perturbations enrich the activation space with orthogonal spectral modes, similar to random Fourier features used for kernel approximation (\cite{rahimi2007random}). This expansion enables the adapter to better capture fine-grained nonlinear variation in downstream gradients that conventional activations like ReLU or Tanh may suppress.

This phenomenon parallels the benefits observed in periodic activation networks such as \cite{sitzmann2020implicit}, where sine waves facilitate the modeling of high-frequency details. In LoRAN, we adapt this insight to the low-rank fine-tuning context, leveraging Sinter as a lightweight spectral augmenter.

\subsubsection{Generalization Bounds of Nonlinear Low-Rank Models} 

Another important theoretical dimension is the generalization behavior of nonlinear low-rank modules. Classical results from statistical learning theory suggest that introducing non-linearities increases hypothesis class capacity (e.g., \cite{vapnik2015uniform} or \cite{bartlett2002rademacher}). However, in the case of LoRAN, the non-linearity is structured and bounded—modulated by fixed constants $A$ and $\omega$—which controls the function class expansion.

We conjecture that this structure imposes a regularized complexity bound: while the hypothesis space grows, the increase in expressiveness is tamed by the smooth, input-sensitive design of the perturbation. As a result, LoRAN can enhance representation power without significantly worsening overfitting risk. Moreover, by preserving LoRA’s frozen backbone and constraining adaptation to a structured subspace, LoRAN maintains favorable implicit regularization effects commonly observed in PEFT methods, while expanding the representational envelope in a controllable manner.

\subsection{Code and Model Availability.}
For the convenience of the community, we open-source the training and inference codes of the \href{https://github.com/konolmyda/loran_checkpoint}{LoRAN} framework. It is built upon the widely-used \href{https://github.com/huggingface/peft/tree/v0.4.0}{PEFT} library from Huggingface, enabling seamless integration into existing PEFT workflows. Models are trained on a single NVIDIA A100 GPU (40GB).

\section{Conclusion and Future Work}

This paper presents \textbf{LoRAN}, a novel extension of Low-Rank Adaptation (LoRA) that introduces non-linear transformations to mitigate the expressiveness bottleneck commonly observed in low-rank parameter-efficient fine-tuning. Through the design of \textbf{Sinter}—a lightweight sine-based activation function—we enable structured, amplitude-controlled perturbations that enrich the LoRA update space without increasing parameter count or compromising training efficiency.

Our experiments demonstrate that LoRAN consistently outperforms strong baselines such as QLoRA across both summarization and classification tasks, especially in low-resource or under-parameterized settings. In addition, Sinter exhibits strong performance compared to standard activations like Swish and Tanh, validating the importance of activation design in enhancing low-rank expressiveness. Extensive stability analyses further show that while non-linear activations offer training benefits, periodic activations like Sinter may require calibration for use in larger models.

Future work will explore several directions. First, we plan to study the theoretical convergence properties of LoRAN in both convex and non-convex settings. Second, while Sinter provides one effective instance of structured non-linearity, other perturbation families—e.g., orthogonal bases, Fourier filters, or learnable interference patterns—may offer further benefits. Finally, we are interested in extending LoRAN beyond transformer-based architectures to lightweight vision backbones or reinforcement learning agents, where training efficiency and robustness are equally critical.

\begin{table}[t]
    \centering
    \fontsize{8}{9.5}\selectfont  
    \bgroup
    \def\arraystretch{1.3}  
    \begin{tabular}{c|ccc}
    \hline
    \multirow{2}{*}{Foundation Model} & \multicolumn{3}{c}{Training Cost (GPU hours)} \\
    & QLoRA & LoRAN & $\Delta$ \\\hline
    LLaMA-2-7B    & 2.9 & 3.1 & +0.2 \\
    LLaMA-2-13B   & 4.9 & 5.4 & +0.5 \\\hline
    \end{tabular}
    \caption{Comparison of the time consumption of QLoRA and LoRAN methods on the 20 Newsgroups task.}
    \label{table:gpu_hours}
    \egroup
\end{table}
\begin{table}[t]
    \centering
    \fontsize{8}{9.5}\selectfont  
    \bgroup
    \def\arraystretch{1.3}  
    \begin{tabular}{ccc}
    \hline
    \multicolumn{2}{c}{Categories} & \multirow{2}{*}{Sample Number} \\
    Broad & Fine-grained & \\\hline
    \multirow{5}{*}{A} & comp.graphics (\#1) & 550 \\
    & comp.os.ms-windows.misc (\#2) & 554 \\
    & comp.sys.ibm.pc.hardware (\#3) & 561 \\
    & comp.sys.mac.hardware (\#4) & 536 \\
    & comp.windows.x (\#5) & 575 \\\hline
    \multirow{4}{*}{B} & rec.autos (\#6) & 538 \\
    & rec.motorcycles (\#7) & 550 \\
    & rec.sport.baseball (\#8) & 546 \\
    & rec.sport.hockey (\#9) & 558 \\\hline
    \multirow{4}{*}{C} & sci.crypt (\#10) & 567 \\
    & sci.electronics (\#11) & 562 \\
    & sci.med (\#12) & 571 \\
    & sci.space (\#13) & 563 \\\hline
    D & misc.forsale (\#14) & 564 \\\hline
    \multirow{3}{*}{E} & talk.politics.misc (\#15) & 437 \\
    & talk.politics.guns (\#16) & 525 \\
    & talk.politics.mideast (\#17) & 520 \\\hline
    \multirow{3}{*}{F} & talk.religion.misc (\#18) & 338 \\
    & alt.atheism (\#19) & 448 \\
    & soc.religion.christian (\#20) & 581 \\\hline
    \end{tabular}
    \caption{Broad and fine-grained categories in the 20 newsgroups
task. Each fine-grained category corresponds to a
specific topic, while similar topics are clustered
into a broad category.}
    \label{table:stats_newsgroups}
    \egroup
\end{table}

\section{Limitations}
While LoRAN demonstrates consistent improvements over existing PEFT methods, two major limitations remain. First, although Sinter is lightweight, its periodic nature may introduce mild training instabilities in large-scale models under certain conditions. Second, While LoRAN introduces no additional parameters, the non-linear transformation does incur a modest increase in computational time. In our experiments with LLaMA-2 (7B and 13B), this overhead remains minimal—under 0.5 GPU hours (see Table~\ref{table:gpu_hours}). Furthermore, such cost can be further reduced with kernel fusion or operator-level optimization techniques.

\section{Acknowledgements}
This work was supported in part by the Research Grants Council of the Hong Kong SAR under Grant GRF 11217823, Grant GRF 11216225 and Collaborative Research Fund C1042-23GF, the National Natural Science Foundation of China under Grant 62371411, InnoHK initiative, the Government of the HKSAR, Laboratory for AI-Powered Financial Technologies.

\begin{figure}[t]
  \centering
  \begin{subfigure}[t]{0.9\linewidth}
    \centering
    \includegraphics[width=\linewidth]{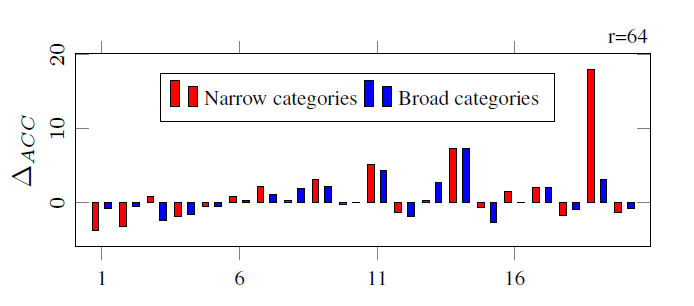}
    \caption{$r = 64$}
    \label{fig:newsgroups_rank64}
  \end{subfigure}
  \vskip 0.5em  
  \begin{subfigure}[t]{0.9\linewidth}
    \centering
    \includegraphics[width=\linewidth]{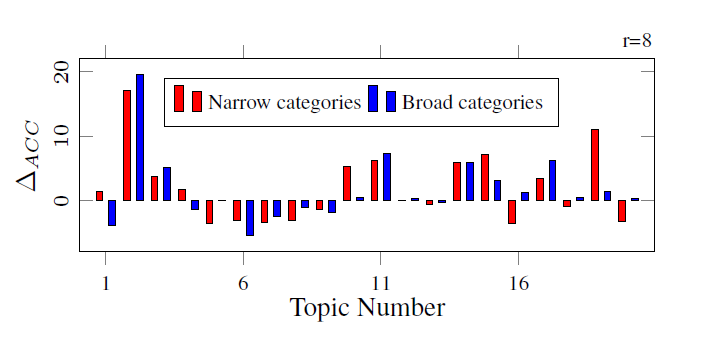}
    \caption{$r = 8$}
    \label{fig:newsgroups_rank8}
  \end{subfigure}
  \caption{Topic-level classification accuracy of QLoRA and LoRAN on the 20 Newsgroups task. The foundation model is Llama-2-7b.}
  \label{fig:newsgroups_rank}
\end{figure}

{\appendix[20 Newsgroups Task]

As shown in Table~\ref{table:stats_newsgroups}, the 20 classes correspond to fine-grained news topics, which are further grouped into five broader categories based on semantic similarity. The table also details the number of samples per class.

Figure~\ref{fig:newsgroups_rank} illustrates the performance improvement of LoRAN over QLoRA for each class. For fine-grained categories (marked in red), a correct prediction requires an exact match with the gold label. In contrast, for broad categories (marked in blue), a prediction is considered correct if it falls within the correct broader group.

When the rank is high ($r = 64$), LoRAN’s performance gains primarily arise from fine-grained categories, as it can leverage sufficient parameters to fit more nuanced patterns. Conversely, at a lower rank ($r = 8$), improvements are mostly seen in broad categories, which are easier to predict and require less representational capacity.

This pattern reflects the relative difficulty of the tasks: distinguishing broad categories is generally easier than identifying specific fine-grained classes. With $r = 64$, LoRA can already achieve high accuracy on broad categories, leaving room for LoRAN to improve upon the harder fine-grained ones through its non-linear transformation mechanism. However, when the rank drops to $r = 8$, LoRA's linear adapter lacks the capacity to reliably separate even broad categories. In such cases, LoRAN’s non-linearity shifts focus toward the simpler task—broad category prediction—resulting in greater gains in those classes, as reflected in Figure~\ref{fig:newsgroups_rank8}. This demonstrates that LoRAN can enhance LoRA’s performance even under constrained settings such as low-rank configurations.
}

\begingroup
\let\clearpage\relax
\let\newpage\relax
\bibliographystyle{IEEEtran}
\bibliography{ref}

\begin{thebibliography}{10}
\providecommand{\url}[1]{#1}
\csname url@samestyle\endcsname
\providecommand{\newblock}{\relax}
\providecommand{\bibinfo}[2]{#2}
\providecommand{\BIBentrySTDinterwordspacing}{\spaceskip=0pt\relax}
\providecommand{\BIBentryALTinterwordstretchfactor}{4}
\providecommand{\BIBentryALTinterwordspacing}{\spaceskip=\fontdimen2\font plus
\BIBentryALTinterwordstretchfactor\fontdimen3\font minus \fontdimen4\font\relax}
\providecommand{\BIBforeignlanguage}[2]{{%
\expandafter\ifx\csname l@#1\endcsname\relax
\typeout{** WARNING: IEEEtran.bst: No hyphenation pattern has been}%
\typeout{** loaded for the language `#1'. Using the pattern for}%
\typeout{** the default language instead.}%
\else
\language=\csname l@#1\endcsname
\fi
#2}}
\providecommand{\BIBdecl}{\relax}
\BIBdecl

\bibitem{brown2020language}
T.~Brown, B.~Mann, N.~Ryder, M.~Subbiah, J.~D. Kaplan, P.~Dhariwal, A.~Neelakantan, P.~Shyam, G.~Sastry, A.~Askell \emph{et~al.}, ``Language models are few-shot learners,'' \emph{Advances in neural information processing systems}, vol.~33, pp. 1877--1901, 2020.

\bibitem{zhao2023survey}
W.~X. Zhao, K.~Zhou, J.~Li, T.~Tang, X.~Wang, Y.~Hou, Y.~Min, B.~Zhang, J.~Zhang, Z.~Dong \emph{et~al.}, ``A survey of large language models,'' \emph{arXiv preprint arXiv:2303.18223}, vol.~1, no.~2, 2023.

\bibitem{raffel2020exploring}
C.~Raffel, N.~Shazeer, A.~Roberts, K.~Lee, S.~Narang, M.~Matena, Y.~Zhou, W.~Li, and P.~J. Liu, ``Exploring the limits of transfer learning with a unified text-to-text transformer,'' \emph{Journal of machine learning research}, vol.~21, no. 140, pp. 1--67, 2020.

\bibitem{hu2022lora}
E.~J. Hu, Y.~Shen, P.~Wallis, Z.~Allen-Zhu, Y.~Li, S.~Wang, L.~Wang, W.~Chen \emph{et~al.}, ``Lora: Low-rank adaptation of large language models.'' \emph{ICLR}, vol.~1, no.~2, p.~3, 2022.

\bibitem{liu2021p}
X.~Liu, K.~Ji, Y.~Fu, W.~L. Tam, Z.~Du, Z.~Yang, and J.~Tang, ``P-tuning v2: Prompt tuning can be comparable to fine-tuning universally across scales and tasks,'' \emph{arXiv preprint arXiv:2110.07602}, 2021.

\bibitem{lester2021power}
B.~Lester, R.~Al-Rfou, and N.~Constant, ``The power of scale for parameter-efficient prompt tuning,'' \emph{arXiv preprint arXiv:2104.08691}, 2021.

\bibitem{aghajanyan2020intrinsic}
A.~Aghajanyan, L.~Zettlemoyer, and S.~Gupta, ``Intrinsic dimensionality explains the effectiveness of language model fine-tuning,'' \emph{arXiv preprint arXiv:2012.13255}, 2020.

\bibitem{dettmers2023qlora}
T.~Dettmers, A.~Pagnoni, A.~Holtzman, and L.~Zettlemoyer, ``Qlora: Efficient finetuning of quantized llms,'' \emph{Advances in neural information processing systems}, vol.~36, pp. 10\,088--10\,115, 2023.

\bibitem{li2024loran}
Y.~Li, L.~Song, and H.~Hou, ``Loran: Improved low-rank adaptation by a non-linear transformation,'' in \emph{Findings of the Association for Computational Linguistics: EMNLP 2024}, 2024, pp. 3134--3143.

\bibitem{houlsby2019parameter}
N.~Houlsby, A.~Giurgiu, S.~Jastrzebski, B.~Morrone, Q.~De~Laroussilhe, A.~Gesmundo, M.~Attariyan, and S.~Gelly, ``Parameter-efficient transfer learning for nlp,'' in \emph{International conference on machine learning}.\hskip 1em plus 0.5em minus 0.4em\relax PMLR, 2019, pp. 2790--2799.

\bibitem{li2021prefix}
X.~L. Li and P.~Liang, ``Prefix-tuning: Optimizing continuous prompts for generation,'' \emph{arXiv preprint arXiv:2101.00190}, 2021.

\bibitem{zhang2023adalora}
Q.~Zhang, M.~Chen, A.~Bukharin, N.~Karampatziakis, P.~He, Y.~Cheng, W.~Chen, and T.~Zhao, ``Adalora: Adaptive budget allocation for parameter-efficient fine-tuning,'' \emph{arXiv preprint arXiv:2303.10512}, 2023.

\bibitem{valipour2022dylora}
M.~Valipour, M.~Rezagholizadeh, I.~Kobyzev, and A.~Ghodsi, ``Dylora: Parameter efficient tuning of pre-trained models using dynamic search-free low-rank adaptation,'' \emph{arXiv preprint arXiv:2210.07558}, 2022.

\bibitem{hyeon2021fedpara}
N.~Hyeon-Woo, M.~Ye-Bin, and T.-H. Oh, ``Fedpara: Low-rank hadamard product for communication-efficient federated learning,'' \emph{arXiv preprint arXiv:2108.06098}, 2021.

\bibitem{kopiczko2023vera}
D.~J. Kopiczko, T.~Blankevoort, and Y.~M. Asano, ``Vera: Vector-based random matrix adaptation,'' \emph{arXiv preprint arXiv:2310.11454}, 2023.

\bibitem{hayou2024lora+}
S.~Hayou, N.~Ghosh, and B.~Yu, ``Lora+: Efficient low rank adaptation of large models,'' \emph{arXiv preprint arXiv:2402.12354}, 2024.

\bibitem{jiang2024mora}
T.~Jiang, S.~Huang, S.~Luo, Z.~Zhang, H.~Huang, F.~Wei, W.~Deng, F.~Sun, Q.~Zhang, D.~Wang \emph{et~al.}, ``Mora: High-rank updating for parameter-efficient fine-tuning,'' \emph{arXiv preprint arXiv:2405.12130}, 2024.

\bibitem{wedin1972perturbation}
P.-{\AA}. Wedin, ``Perturbation bounds in connection with singular value decomposition,'' \emph{BIT Numerical Mathematics}, vol.~12, pp. 99--111, 1972.

\bibitem{proakis2008digital}
J.~G. Proakis and M.~Salehi, \emph{Digital communications}.\hskip 1em plus 0.5em minus 0.4em\relax McGraw-hill, 2008.

\bibitem{hornik1991approximation}
K.~Hornik, ``Approximation capabilities of multilayer feedforward networks,'' \emph{Neural networks}, vol.~4, no.~2, pp. 251--257, 1991.

\bibitem{he2016deep}
K.~He, X.~Zhang, S.~Ren, and J.~Sun, ``Deep residual learning for image recognition,'' in \emph{Proceedings of the IEEE conference on computer vision and pattern recognition}, 2016, pp. 770--778.

\bibitem{zhong2024neat}
Y.~Zhong, H.~Jiang, L.~Li, R.~Nakada, T.~Liu, L.~Zhang, H.~Yao, and H.~Wang, ``Neat: Nonlinear parameter-efficient adaptation of pre-trained models,'' \emph{arXiv preprint arXiv:2410.01870}, 2024.

\bibitem{dong2025aurora}
H.~Dong, W.~Zhu, G.~Song, and L.~Wang, ``Aurora: Breaking low-rank bottleneck of lora with nonlinear mapping,'' \emph{arXiv preprint arXiv:2505.18738}, 2025.

\bibitem{glorot2010understanding}
X.~Glorot and Y.~Bengio, ``Understanding the difficulty of training deep feedforward neural networks,'' in \emph{Proceedings of the thirteenth international conference on artificial intelligence and statistics}.\hskip 1em plus 0.5em minus 0.4em\relax JMLR Workshop and Conference Proceedings, 2010, pp. 249--256.

\bibitem{nair2010rectified}
V.~Nair and G.~E. Hinton, ``Rectified linear units improve restricted boltzmann machines,'' in \emph{Proceedings of the 27th international conference on machine learning (ICML-10)}, 2010, pp. 807--814.

\bibitem{maas2013rectifier}
A.~L. Maas, A.~Y. Hannun, A.~Y. Ng \emph{et~al.}, ``Rectifier nonlinearities improve neural network acoustic models,'' in \emph{Proc. icml}, vol.~30, no.~1.\hskip 1em plus 0.5em minus 0.4em\relax Atlanta, GA, 2013, p.~3.

\bibitem{he2015delving}
K.~He, X.~Zhang, S.~Ren, and J.~Sun, ``Delving deep into rectifiers: Surpassing human-level performance on imagenet classification,'' in \emph{Proceedings of the IEEE international conference on computer vision}, 2015, pp. 1026--1034.

\bibitem{clevert2015fast}
D.-A. Clevert, T.~Unterthiner, and S.~Hochreiter, ``Fast and accurate deep network learning by exponential linear units (elus),'' \emph{arXiv preprint arXiv:1511.07289}, 2015.

\bibitem{ramachandran2017searching}
P.~Ramachandran, B.~Zoph, and Q.~V. Le, ``Searching for activation functions,'' \emph{arXiv preprint arXiv:1710.05941}, 2017.

\bibitem{zhou2024lora}
H.~Zhou, X.~Lu, W.~Xu, C.~Zhu, T.~Zhao, and M.~Yang, ``Lora-drop: Efficient lora parameter pruning based on output evaluation,'' \emph{arXiv preprint arXiv:2402.07721}, 2024.

\bibitem{yeh2023navigating}
S.-Y. Yeh, Y.-G. Hsieh, Z.~Gao, B.~B. Yang, G.~Oh, and Y.~Gong, ``Navigating text-to-image customization: From lycoris fine-tuning to model evaluation,'' in \emph{The Twelfth International Conference on Learning Representations}, 2023.

\bibitem{zhang2023lora}
L.~Zhang, L.~Zhang, S.~Shi, X.~Chu, and B.~Li, ``Lora-fa: Memory-efficient low-rank adaptation for large language models fine-tuning,'' \emph{arXiv preprint arXiv:2308.03303}, 2023.

\bibitem{gliwa2019samsum}
B.~Gliwa, I.~Mochol, M.~Biesek, and A.~Wawer, ``Samsum corpus: A human-annotated dialogue dataset for abstractive summarization,'' \emph{arXiv preprint arXiv:1911.12237}, 2019.

\bibitem{lang1995newsweeder}
K.~Lang, ``Newsweeder: Learning to filter netnews,'' in \emph{Machine learning proceedings 1995}.\hskip 1em plus 0.5em minus 0.4em\relax Elsevier, 1995, pp. 331--339.

\bibitem{chung2024scaling}
H.~W. Chung, L.~Hou, S.~Longpre, B.~Zoph, Y.~Tay, W.~Fedus, Y.~Li, X.~Wang, M.~Dehghani, S.~Brahma \emph{et~al.}, ``Scaling instruction-finetuned language models,'' \emph{Journal of Machine Learning Research}, vol.~25, no.~70, pp. 1--53, 2024.

\bibitem{penedo2023refinedweb}
G.~Penedo, Q.~Malartic, D.~Hesslow, R.~Cojocaru, A.~Cappelli, H.~Alobeidli, B.~Pannier, E.~Almazrouei, and J.~Launay, ``The refinedweb dataset for falcon llm: outperforming curated corpora with web data, and web data only,'' \emph{arXiv preprint arXiv:2306.01116}, 2023.

\bibitem{touvron2023llama}
H.~Touvron, L.~Martin, K.~Stone, P.~Albert, A.~Almahairi, Y.~Babaei, N.~Bashlykov, S.~Batra, P.~Bhargava, S.~Bhosale \emph{et~al.}, ``Llama 2: Open foundation and fine-tuned chat models,'' \emph{arXiv preprint arXiv:2307.09288}, 2023.

\bibitem{srivastava2014dropout}
N.~Srivastava, G.~Hinton, A.~Krizhevsky, I.~Sutskever, and R.~Salakhutdinov, ``Dropout: a simple way to prevent neural networks from overfitting,'' \emph{The journal of machine learning research}, vol.~15, no.~1, pp. 1929--1958, 2014.

\bibitem{tishby2000information}
N.~Tishby, F.~C. Pereira, and W.~Bialek, ``The information bottleneck method,'' \emph{arXiv preprint physics/0004057}, 2000.

\bibitem{shwartz2017opening}
R.~Shwartz-Ziv and N.~Tishby, ``Opening the black box of deep neural networks via information,'' \emph{arXiv preprint arXiv:1703.00810}, 2017.

\bibitem{rahimi2007random}
A.~Rahimi and B.~Recht, ``Random features for large-scale kernel machines,'' \emph{Advances in neural information processing systems}, vol.~20, 2007.

\bibitem{sitzmann2020implicit}
V.~Sitzmann, J.~Martel, A.~Bergman, D.~Lindell, and G.~Wetzstein, ``Implicit neural representations with periodic activation functions,'' \emph{Advances in neural information processing systems}, vol.~33, pp. 7462--7473, 2020.

\bibitem{vapnik2015uniform}
V.~N. Vapnik and A.~Y. Chervonenkis, ``On the uniform convergence of relative frequencies of events to their probabilities,'' in \emph{Measures of complexity: festschrift for alexey chervonenkis}.\hskip 1em plus 0.5em minus 0.4em\relax Springer, 2015, pp. 11--30.

\bibitem{bartlett2002rademacher}
P.~L. Bartlett and S.~Mendelson, ``Rademacher and gaussian complexities: Risk bounds and structural results,'' \emph{Journal of Machine Learning Research}, vol.~3, no. Nov, pp. 463--482, 2002.

\end{thebibliography}
\endgroup

\end{document}